\documentclass[11pt]{article}

\usepackage[final]{acl}

\usepackage{times}
\usepackage{latexsym}
\usepackage{etoolbox}

\usepackage[T1]{fontenc}

\usepackage[utf8]{inputenc}
\usepackage{amsmath}
\usepackage{amsfonts}
\usepackage{amssymb}
\usepackage{multirow}
\usepackage[most]{tcolorbox}
\usepackage{enumitem}
\usepackage{ragged2e}
\usepackage{tabularx}
\usepackage{booktabs}
\usepackage{float}

\usepackage{booktabs}
\usepackage[table]{xcolor}   
\usepackage{siunitx}         

\usepackage{microtype}

\usepackage{inconsolata}

\usepackage{graphicx}

\NewDocumentCommand{\hongru}
{ mO{} }{\textcolor{blue}{\textsuperscript{\textit{Hongru}}\textsf{\textbf{\small[#1]}}}}

\title{%
  \raisebox{1cm}[0pt][0pt]{%
    \makebox[\textwidth][c]{%
      \scalebox{1}[1]{%
        \includegraphics[height=0.88cm,trim=0 4 0 4,clip]{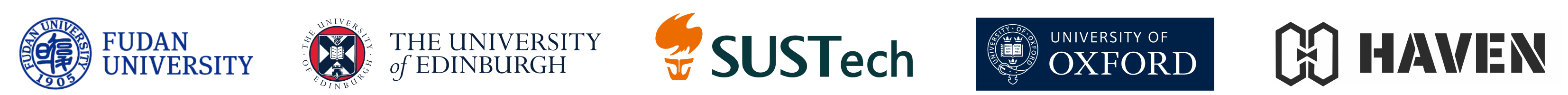}%
      }%
    }%
  }\\[-2.8em]
  \rule{\textwidth}{0.4pt}\\[1.2em]
  Rethinking the Role of Entropy in Optimizing Tool-Use Behaviors for Large Language Model Agents
}

\author{
  Zeping Li$^{1}$,
  Hongru Wang$^{2,\dagger}$,
  Yiwen Zhao$^{1}$,
  Guanhua Chen$^{3}$,
  Yixia Li$^{3}$,
  Keyang Chen$^{1}$, \\
  \textbf{Yixin Cao}$^{1}$,
  \textbf{Guangnan Ye}$^{1}$,
  \textbf{Hongfeng Chai}$^{1,\dagger}$,
  \textbf{Zhenfei Yin}$^{4,5,\dagger}$ \\
  $^{1}$Fudan University \quad
  $^{2}$University of Edinburgh \quad
  $^{3}$Southern University of Science and Technology \quad \\
  $^{4}$Oxford University 
  $^{5}$Haven
  $^{\dagger}$Corresponding authors
}

\begin{document}
\maketitle
\begin{abstract}
Tool-using agents based on Large Language Models (LLMs) excel in tasks such as mathematical reasoning and multi-hop question answering. However, in long trajectories, agents often trigger excessive and low-quality tool calls, increasing latency and degrading inference performance, making managing tool-use behavior challenging. In this work, we conduct entropy-based pilot experiments and observe a strong positive correlation between entropy reduction and high-quality tool calls. Building on this finding, we propose using entropy reduction as a supervisory signal and design two reward strategies to address the differing needs of optimizing tool-use behavior. Sparse outcome rewards provide coarse, trajectory-level guidance to improve efficiency, while dense process rewards offer fine-grained supervision to enhance performance. Experiments across diverse domains show that both reward designs improve tool-use behavior: the former reduces tool calls by 72.07\% compared to the average of baselines, while the latter improves performance by 22.27\%. These results position entropy reduction as a key mechanism for enhancing tool-use behavior, enabling agents to be more adaptive in real-world applications.
\end{abstract}

\section{Introduction}
Large language models are increasingly deployed as tool-using agents, where solving a task requires interleaving natural language reasoning with external actions such as code execution, web search, or API calls~\cite{qian2025toolrl, li2025torl, jin2025searchr1trainingllmsreason}. This agentic paradigm has demonstrated strong performance on complex reasoning benchmarks, including mathematical problem solving~\cite{cobbe2021trainingverifierssolvemath}, multi-hop question answering~\cite{yang2018hotpotqa, ho2020constructing, trivedi2022musique}, and deep searching tasks~\cite{mialon2023gaia, wu2025webwalker, phan2025humanity}. 

However, in long trajectories, agents may invoke tools excessively or inappropriately, increasing computation cost and derailing the reasoning process~\cite{qian2025smart}. Therefore, effective tool use poses a new challenge: an agent must not only produce a correct final answer, but also decide when and how to invoke tools throughout a long-horizon reasoning process~\cite{zhang2025rlvmr}, highlighting the importance of tool-invocation behavior itself~\cite{wang2025toward}.

\begin{figure}[t!]
  \centering
  \includegraphics[width=\columnwidth]{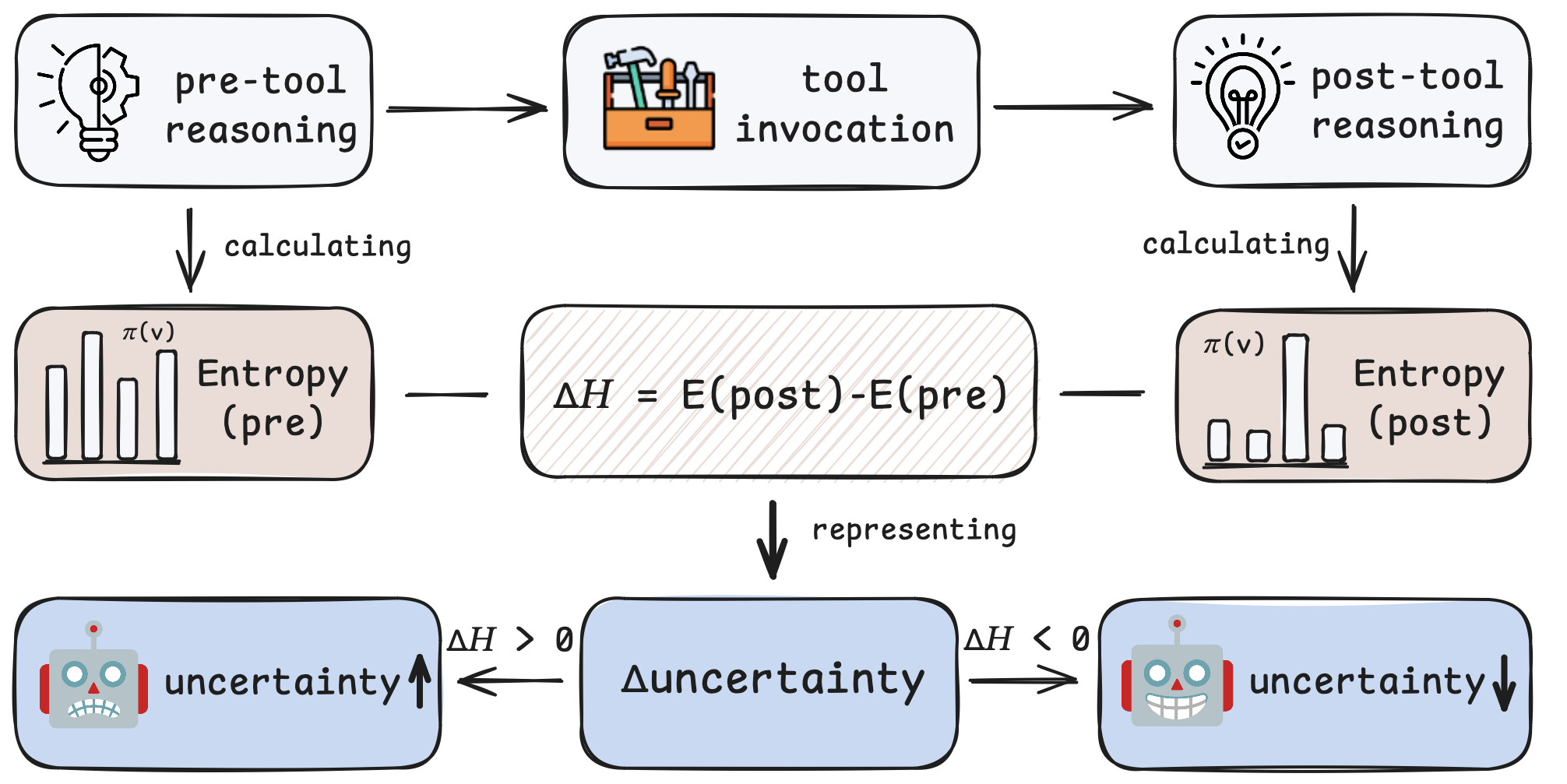}
  \caption{Changes in entropy reflect shifts in uncertainty within the agent. High-quality tool calls help the model reduce uncertainty, as indicated by a decrease in entropy.}
  \label{fig:intro_case}
\end{figure}

Most recent existing works adopt final-answer correctness as the outcome reward~\cite{gao2025beyond, zhang2025tool}: while this supervision can teach models to use tools, it provides only end-point feedback and thus fails to make the agent attend to tool-invocation behavior itself. In contrast, recent process-reward reinforcement learning (RL) methods~\cite{feng2025group, zhang2025rlvmr}, which focus on interactive environments like navigation tasks or object manipulation, deliver fine-grained supervision at each tool-use step, better guiding agents toward more accurate tool usage. Nonetheless, these process-reward designs typically rely on externally specified notions of ``good'' intermediate behaviors (e.g., hand-crafted rules, structured annotations), which are difficult to design for more complex long-term reasoning tasks. Consequently, a principled and lightweight signal for evaluating tool-call quality during long-horizon reasoning remains underexplored.

In this work, we revisit model uncertainty as a source of supervision for tool-augmented reasoning~\cite{li2025confidence, zhao2025learning, cheng2025reasoning}. Through entropy-based pilot experiments on tool-using models, we observe a consistent pattern across multiple domains: high-quality tool calls are generally followed by a reduction in entropy, whereas low-quality tool calls tend to increase it. This can be understood through cognitive load theory~\cite{sweller2011cognitive}: the quality of a tool call depends on whether it reduces unnecessary processing or introduces irrelevant information, thus affecting the model's uncertainty. From this, we derive a general insight: the change in entropy before and after a tool call, quantified as delta segment entropy, serves as an intrinsic signal that reflects tool effectiveness, independent of task-specific annotations or external evaluators.

Building on this insight, we use delta segment entropy as a lightweight supervisory signal. Accordingly, we propose \textbf{T}ool-enhanced \textbf{E}ntropy-guided \textbf{P}olicy \textbf{O}ptimization (TEPO), introducing two reward designs to address different aspects of optimizing tool-use behavior. Specifically, $\text{TEPO}_{\text{sparse}}$ focuses on improving tool-use efficiency by applying a sparse outcome reward that modulates the final-task reward based on the proportion of entropy-decreasing tool calls; whereas $\text{TEPO}_{\text{dense}}$ emphasizes performance gains, using a dense process reward that grants an additional bonus whenever a tool call reduces entropy, thereby providing fine-grained supervision. Extensive experiments across domains demonstrate the effectiveness of both designs. $\text{TEPO}_{\text{sparse}}$ improves efficiency, reducing tool calls by 72.07\% with comparable results, whereas $\text{TEPO}_{\text{dense}}$ focuses on boosting overall performance with an 22.27\% improvement. 

In summary, main contributions are as follows:

\begin{itemize}
    \item We empirically observe that high-quality tool calls often induce entropy reduction, yielding a model-agnostic signal of tool-call quality across diverse reasoning domains.

    \item The proposed reward designs, $\text{TEPO}_{\text{sparse}}$ and $\text{TEPO}_{\text{dense}}$, enhance tool behavior with distinct focuses on efficiency or performance, addressing the diverse needs across tasks.

    \item Comprehensive experiments demonstrate that entropy reduction is a crucial factor in optimizing tool-use behavior, empowering agents to adapt more effectively to real-world tasks.

\end{itemize}

\section{Preliminary}
\subsection{Formalization of Agents' Tool Use}
We formalize tool-augmented generation as an iterative interaction between an agent $A$ and a tool executor $E$. 
Given a query $Q$, the system state at step $k$ is defined as
\begin{equation}
    s_k = (r_1, a_1, o_1, \dots, r_k, a_k, o_k),
\end{equation}
where $r_i$ denotes the natural-language reasoning generated at step $i$, $a_i$ denotes the tool call issued at step $i$, and $o_i$ denotes the observation returned by the executor $E$ after executing $a_i$. 

At each step, the agent conditions on $Q$ and the interaction history $s_{k-1}$ to produce the next reasoning segment and tool call, and the executor returns the corresponding observation. The interaction terminates when the agent emits a designated \emph{finish} action, yielding the final answer.

\subsection{Formalization of Delta Segment Entropy}
\label{formalization of delta entropy}
At each decoding step $t$, the agent model (parameterized by $\theta$) produces a hidden representation $h_t$, which yields a next-token distribution $\pi_\theta(\cdot \mid h_t)$ over the vocabulary $V$. We define the token-level entropy as:
\begin{equation}
    H(h_t) = - \sum_{v \in V} \pi_\theta(v \mid h_t) \log \pi_\theta(v \mid h_t),
\end{equation}
which measures the uncertainty of the model's next-token prediction at step \( t \).

Given a reasoning segment $r$ consisting of a contiguous sequence of tokens, we define the segment-level entropy as the token-entropy averaged over all tokens within the segment:
\begin{equation}
    H(r) = \frac{1}{|I|} \sum_{t \in I} H(h_t),
\end{equation}
where $I$ denotes the index set of tokens belonging to the reasoning segment $r$. This formulation captures the overall uncertainty exhibited by the model throughout the reasoning process.

\begin{table}[t]
\centering
\small
\sisetup{detect-all, table-number-alignment=center}

\resizebox{\columnwidth}{!}{
\begin{tabular}{@{\extracolsep{\fill}} l ccccc}
\toprule
\textbf{Entropy} & \textbf{Score} & \textbf{Math} & \textbf{Search} & \textbf{DeepS.} & \textbf{Avg} \\

\midrule
\rowcolor{gray!12}
\multicolumn{6}{c}{\textbf{\textit{Qwen3-8B-SFT}}} \\
\midrule

$\Delta H_k$                & 0 & 17.81  & 37.90  & 2.72   & 19.47 \\
$\Delta H_k$                & 1 & -37.27 & -14.48 & -23.07 & \textbf{-24.94} \\
\cmidrule(lr){1-6}
$\Delta H_k^{\text{ratio}}$ & 0 & 48.62  & 18.77  & 16.14  & 27.51 \\
$\Delta H_k^{\text{ratio}}$ & 1 & 50.89  & 26.77  & 20.47  & 32.04 \\

\midrule
\rowcolor{gray!12}
\multicolumn{6}{c}{\textbf{\textit{Llama3.1-8B-SFT}}} \\
\midrule
$\Delta H_k$                & 0 & -0.96  & 32.95  & 2.08   & 11.35 \\
$\Delta H_k$                & 1 & -41.03 & 1.38   & -30.29 & \textbf{-23.64} \\
\cmidrule(lr){1-6}
$\Delta H_k^{\text{ratio}}$ & 0 & 47.84  & 19.92  & 21.25  & 29.34 \\
$\Delta H_k^{\text{ratio}}$ & 1 & 42.98  & 26.54  & 25.72  & 31.41 \\
\bottomrule
\end{tabular}
}
\caption{Entropy-based pilot experiment results: Average delta segment entropy ($\Delta H_k$) and delta segment entropy ratio ($\Delta H_k^{\text{ratio}}$) of SFT models, grouped by tool score (0/1) across the Math, Search, DeepSearch(DeepS.) domains. For readability, $\Delta H_k$ values are scaled by a factor of $10^3$. }
\label{pilot experiments}
\vspace{-10pt}
\end{table}

Let $r_{k-1}$ denote the reasoning segment generated immediately before the $k$-th tool call, and let $r_k$ denote the subsequent reasoning segment generated after receiving the tool observation $o_k$.
We define the delta segment entropy $\Delta H_k$ between these two segments as:
\begin{equation}
    \Delta H_k = H(r_k) - H(r_{k-1}).
\end{equation}

When $\Delta H_k$ is negative, indicating an entropy reduction, we define the delta segment entropy ratio $\Delta H_k^{\text{ratio}}$ to further quantify the extent of this reduction and enable comparison across different queries and reasoning contexts as:

\begin{equation}
    \Delta H_k^{\text{ratio}} = 
    \frac{H(r_k) - H(r_{k-1})}{H(r_{k-1}) + \epsilon},
\end{equation}
where $\epsilon = 10^{-8}$ is a small constant introduced for numerical stability and to avoid division by zero.

\subsection{Entropy-Based Pilot Experiments}
\label{pilot experiment}
To obtain an initial understanding of how entropy dynamics relate to tool usage, 
we conduct a set of entropy-based pilot experiments. Following the ARPO/AEPO setup~\cite{dong2025arpo,dong2025aepo}, we use the same supervised fine-tuning dataset to train two models, resulting in Qwen3-8B-SFT and Llama3.1-8B-SFT. These SFT-trained models already exhibit basic tool-use capabilities, providing a suitable starting point for our entropy analysis.

We evaluate the SFT models on three domains, following the same domain partition as in ARPO~\cite{dong2025arpo} and AEPO~\cite{dong2025aepo}. To quantify the relationship between tool-call quality and entropy dynamics, we score each tool call with an LLM-as-judge and compute per-call entropy changes after the tool interaction. We then compare entropy statistics between high-quality and low-quality calls (the evaluation details can be found in Appendix~\ref{appendix:A})

The results shown in Table~\ref{pilot experiments} reveal a clear association across all three domains and both SFT models: high-quality tool calls (score~1) consistently yield negative $\Delta H_k$, indicating reduced uncertainty in subsequent reasoning, whereas low-quality calls (score~0) often increase entropy. Moreover, in the Search and DeepSearch domains, score~1 calls exhibit higher $\Delta H_k^{\text{ratio}}$ than score~0 calls. Overall, these findings suggest that entropy reduction serves as a useful, model-agnostic signal correlated with tool effectiveness.

\section{Method}
In this section, we first reformulate GRPO from a token-level perspective to clarify reward attribution to generated tokens(Section~\ref{grpo_perspective}). This unified objective underlies our two reward designs: Section~\ref{sparse_design} applies a sparse outcome reward, while Section~\ref{dense_design} introduces a dense process reward.

\begin{figure*}[t]
  \centering
  \includegraphics[width=\textwidth]{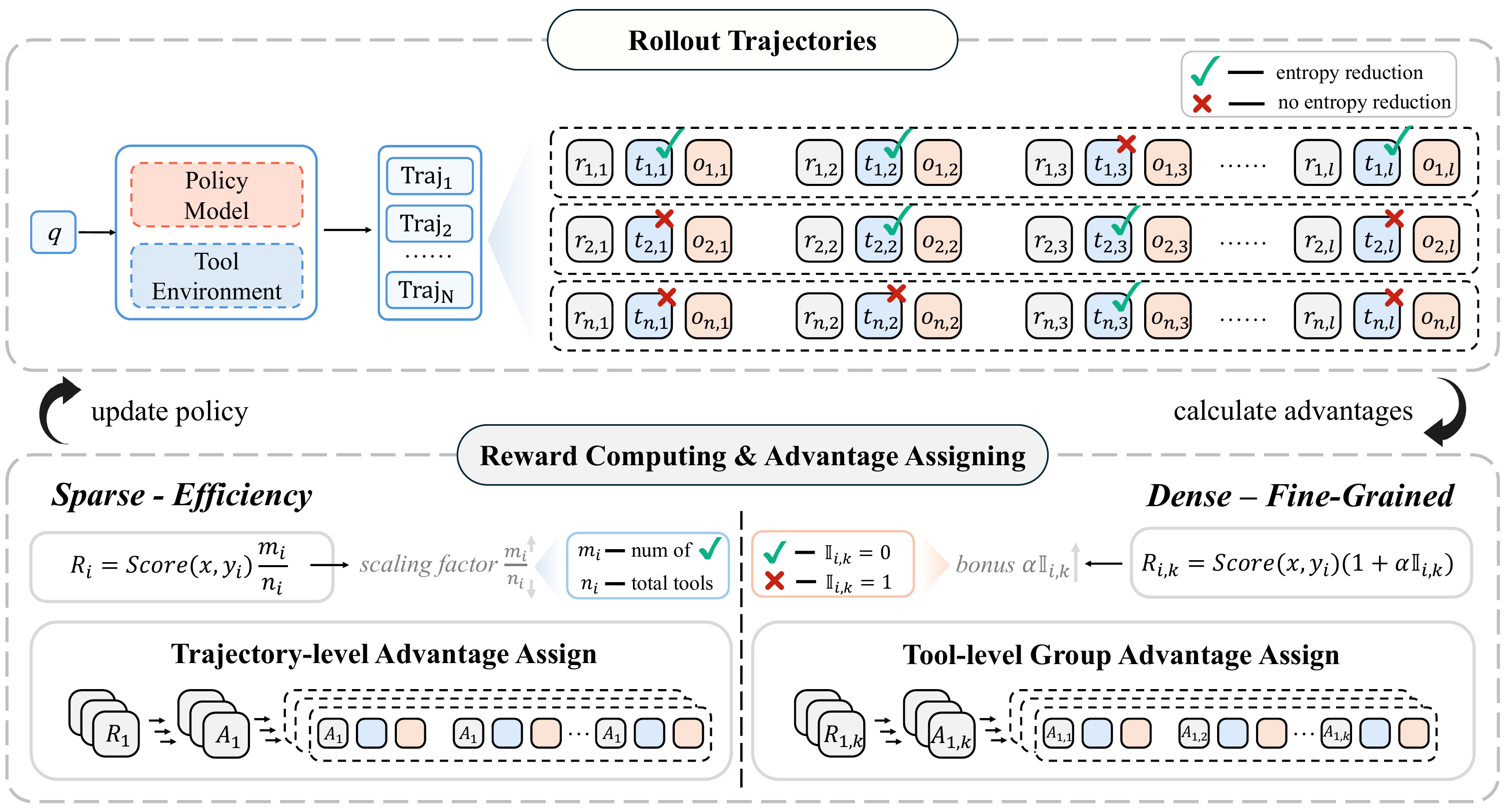}
  \caption{The overall framework of $\text{TEPO}_{\text{sparse}}$ and $\text{TEPO}_{\text{dense}}$. In the sparse reward design, the reward and advantage are calculated and then uniformly assigned to each token within the trajectory (same \( A_{i,t} \) for all tokens). In contrast, the dense reward design assigns fine-grained tool rewards and advantages, resulting in different \( A_{i,t} \) values for different tokens within the same trajectory.}
  \label{fig:main}
\end{figure*}

\subsection{Token-Level GRPO Perspective}
\label{grpo_perspective}
For each input question $x$, we sample a group of $N$ rollouts $\{y_i\}_{i=1}^{N}$ under the GRPO framework. 
Let $y_{i,t}$ denote the token generated at position $t$ in rollout $i$. 
The token-level GRPO objective is formulated as
\begin{equation}
\begin{aligned}
\max_{\theta}\;\mathbb{E}_{x\sim\mathcal{D}}
\Bigg[
\frac{1}{N}\sum_{i=1}^{N}\sum_{t}
\rho_{i,t}(\theta)\,A_{i,t}
\Bigg] \\
\qquad
-\;\beta\,D_{\mathrm{KL}}\!\left(\pi_{\theta}\,\|\,\pi_{\mathrm{ref}}\right),
\end{aligned}
\end{equation}

\noindent where $A_{i,t}$ denotes the advantage assigned to the $t$-th token in rollout $i$, and the token-level importance ratio is defined as
\begin{equation}
\rho_{i,t}(\theta)
=
\frac{\pi_{\theta}(y_{i,t}\mid y_{i,<t},x)}
{\pi_{\theta_{\mathrm{old}}}(y_{i,t}\mid y_{i,<t},x)}.
\end{equation}

\subsection{Sparse Outcome Reward Design}
\label{sparse_design}
Building on the findings from the entropy-based pilot experiments, we adopt a straightforward approach by incorporating the proportion of entropy-decreasing tool calls within a rollout into the reward design. This design encourages the model to either increase the number of entropy-decreasing tool calls or reduce the total number of tool calls to achieve a higher reward. Let $n_i$ denote the total number of tool calls in rollout $i$, and let $m_i$ denote the number of tool calls that induce entropy decrease. We adopt the final-answer F1-score as the verifiable correctness signal, consistent with ARPO~\cite{dong2025arpo} and AEPO~\cite{dong2025aepo}. When a rollout makes no tool calls (i.e., $n_i=0$), the outcome reward reduces to the F1-score only. Otherwise, we define the sparse outcome reward for rollout $i$ as:

\begin{equation}
\label{eq:sparse_reward}
r^{\mathrm{sparse}}_i
=
\mathrm{F1}(x,y_i)\cdot \frac{m_i}{n_i}.
\end{equation}

Following GRPO, the trajectory-level advantage can be calculated as:
\begin{equation}
A^{\mathrm{traj}}_i=\frac{r^{\mathrm{sparse}}_i-\mu^{\mathrm{sparse}}}{\sigma^{\mathrm{sparse}}+\epsilon}.
\end{equation}

As a sparse outcome reward design, we assign the resulting trajectory-level advantage uniformly to all tokens in the rollout. This yields a token-wise assignment $A_{i,t}=A^{\mathrm{traj}}_i$ for every token position $t$ in rollout $i$, providing a stable but coarse-grained credit signal over the entire trajectory.

\subsection{Dense Process Reward Design}
\label{dense_design}
While outcome rewards provide an effective learning signal in reinforcement learning, incorporating process rewards can offer denser supervision and better guide the optimization process. However, obtaining reliable process-level feedback is often challenging for tool-augmented generation. Motivated by the entropy-based pilot experiments, we leverage entropy-decrease signals as a lightweight yet informative proxy for process reward. For the $k$-th tool call in rollout $i$, the tool-level reward can be calculated as:

\begin{equation}
\label{eq:tool_reward}
r^{\mathrm{tool}}_{i,k}
=
\mathrm{F1}(x,y_i)\cdot \left(1+\alpha\,\mathbb{I}_{i,k}\right),
\end{equation}

\noindent where $\mathbb{I}_{i,k}\in\{0,1\}$ is an entropy-decrease indicator, which equals $1$ if the $k$-th tool call in rollout $i$ lowers entropy, and $0$ otherwise. This formulation anchors each tool call to task correctness while explicitly rewarding those that reduce uncertainty. 

\begin{table*}[htbp]
\fontsize{7pt}{12pt}\selectfont
\centering
\caption{Overall performance on reasoning domain tasks. The best results are in \textbf{bold}, and the second-best results are \underline{underlined}. \textbf{$\overline{\text{acc}}$} represents the average accuracy, and \textbf{$\overline{\text{tools}}$} represents the average number of tool calls, both calculated across all five datasets.}
\label{math and search results}
\begin{tabularx}{\textwidth}{l *{7}{>{\centering\arraybackslash}X}}
\toprule
\textbf{Model} & \textbf{aime24} & \textbf{aime25} & \textbf{HotPotQA} & \textbf{2wiki} & \textbf{musique} & \textbf{$\overline{\text{acc}}$} & \textbf{$\overline{\text{tools}}$} \\
\midrule
\textbf{Qwen2.5-7B-Instruct} & 8.67 ± 3.40 & 13.33 ± 4.71 & 47.70 ± 3.31 & 38.50 ± 2.81 & 19.20 ± 1.47 & 25.48 & 1.98 \\
\textbf{\hspace{0.8em}+ SFT} & \underline{24.67 ± 3.40} & 15.33 ± 4.00 & 59.60 ± 1.59 & 56.20 ± 0.81 & 30.20 ± 1.81 & 37.20 & 3.98 \\
\textbf{\hspace{0.8em}+ GRPO} & 23.33 ± 2.98 & 20.00 ± 2.11 & \underline{60.30 ± 0.81} & \textbf{59.30 ± 0.68} & \underline{31.70 ± 1.33} & \underline{38.93} & 1.97 \\
\textbf{\hspace{0.8em}+ ARPO} & 23.33 ± 2.72 & \underline{24.44 ± 1.57} & 56.83 ± 1.55 & 55.17 ± 0.85 & 29.83 ± 1.65 & 37.92 & \underline{1.67} \\
\textbf{\hspace{0.8em}+ AEPO} & 23.33 ± 4.71 & 22.67 ± 6.80 & 58.30 ± 1.12 & 54.20 ± 1.63 & 31.00 ± 1.67 & 37.90 & 2.96 \\
\textbf{\hspace{0.8em}+ $\text{TEPO}_{\text{sparse}}$} & 20.00 ± 2.98 & 21.33 ± 1.63 & 60.10 ± 1.02 & \underline{57.50 ± 1.22} & 29.90 ± 1.36 & 37.77 & \textbf{1.64} \\
\textbf{\hspace{0.8em}+ $\text{TEPO}_{\text{dense}}$} & \textbf{30.67 ± 4.42} & \textbf{30.00 ± 3.65} & \textbf{62.10 ± 1.77} & 57.00 ± 1.00 & \textbf{32.20 ± 1.78} & \textbf{42.39} & 3.99 \\
\midrule
\textbf{Llama3.1-8B-Instruct} & 4.67 ± 2.67 & 2.00 ± 1.63 & 46.70 ± 0.68 & 25.60 ± 2.87 & 17.30 ± 1.44 & 19.25 & 3.45 \\
\textbf{\hspace{0.8em}+ SFT} & 8.67 ± 2.67 & 7.33 ± 3.27 & 59.40 ± 1.32 & 54.30 ± 0.98 & 29.00 ± 1.05 & 31.74 & 5.42 \\
\textbf{\hspace{0.8em}+ GRPO} & 13.33 ± 3.65 & 11.33 ± 1.63 & 62.70 ± 0.68 & \textbf{62.50 ± 0.45} & \underline{32.10 ± 0.80} & 36.39 & 2.02 \\
\textbf{\hspace{0.8em}+ ARPO} & \textbf{20.00 ± 2.98} & \underline{13.33 ± 3.65} & \underline{63.20 ± 0.87} & 61.10 ± 1.66 & 29.20 ± 1.91 & \underline{37.37} & \underline{1.79} \\
\textbf{\hspace{0.8em}+ AEPO} & - & - & - & - & - & - & - \\
\textbf{\hspace{0.8em}+ $\text{TEPO}_{\text{sparse}}$} & \underline{18.67 ± 4.00} & 12.00 ± 1.63 & 60.60 ± 0.58 & 60.70 ± 1.63 & 30.20 ± 0.68 & 36.43 & \textbf{1.66} \\
\textbf{\hspace{0.8em}+ $\text{TEPO}_{\text{dense}}$} & \underline{18.67 ± 3.40} & \textbf{16.67 ± 0.00} & \textbf{64.60 ± 1.16} & \underline{61.70 ± 0.93} & \textbf{33.40 ± 2.84} & \textbf{39.01} & 3.60 \\
\bottomrule
\end{tabularx}
\end{table*}

Since the tools used across the $N$ rollouts of the same question are often similar in type and function, we group these tool rewards and compute tool-level advantages as:

\begin{equation}
\label{eq:tool_adv}
A^{\mathrm{tool}}_{i,k}
=
\frac{
r^{\mathrm{tool}}_{i,k}-\mu_{\mathcal{R}(x)}
}{
\sigma_{\mathcal{R}(x)}+\epsilon
}.
\end{equation}

Here, $\mathcal{R}(x)$ denotes the collection of all tool rewards for question $x$:
\begin{equation}
\label{eq:tool_pool}
\mathcal{R}(x)=\left\{
r^{\mathrm{tool}}_{i,k}\ \middle|\ i\in[1,N],\ k\in[1,n_i]
\right\}.
\end{equation}

We assign token-wise advantages by propagating each tool-level advantage to the reasoning segment before the tool call:
\begin{equation}
\label{eq:dense_token_adv}
A_{i,t}
=
A^{\mathrm{tool}}_{i,k},
\quad \forall t \in I^{\mathrm{pre}}_{i,k},
\end{equation}
where $I^{\mathrm{pre}}_{i,k}$ denotes the token indices of the reasoning segment before the $k$-th tool in rollout $i$.

As a result, different token spans within the same trajectory receive distinct advantages, providing targeted signals that guide the agent to internalize high-quality tool usage and optimize its behavior.

\section{Experiment}
\subsection{Datasets}
To evaluate the effectiveness of our algorithm for training LLM-based tool-using agents, we follow the domain partition adopted in ARPO~\cite{dong2025arpo} and AEPO~\cite{dong2025aepo}, testing on three domains: mathematical reasoning, knowledge-intensive reasoning, and deep information searching. For mathematical reasoning, we use AIME2024 and AIME2025. For knowledge-intensive reasoning, we utilize HotpotQA~\cite{yang2018hotpotqa}, 2WikiMultihopQA~\cite{ho2020constructing}, and Musique~\cite{trivedi2022musique}. For deep information searching, we employ GAIA~\cite{mialon2023gaia}, WebWalker~\cite{wu2025webwalker}, and HLE~\cite{phan2025humanity}. Detailed descriptions of the datasets can be found in the Appendix~\ref{appendix:B.1}.

\subsection{Experimental Settings}
We adopt a two-stage training paradigm, consisting of the SFT stage followed by the RL stage, identical to ARPO/AEPO. This approach not only stabilizes the early-stage optimization process~\cite{dong2025toolstar} but also ensures a fair comparison. All models undergo the same SFT training phase, and for the RL stage, reasoning tasks are primarily tested on Qwen2.5 and Llama3.1 models, while deep search tasks are tested on the Qwen3 model. Details of the SFT and RL training datasets, together with the training parameters and additional information, can be found in Appendix~\ref{implementation details}.

During training and testing, we primarily use two external tools. For computation, we integrate a Python compiler in a sandbox environment that allows safe execution of generated code for complex computation. For knowledge retrieval, we adopt a search setup inspired by the Search‑R1~\cite{jin2025searchr1trainingllmsreason} framework, where the model generates search queries during reasoning and retrieves relevant information from a wiki‑18 corpus.

\begin{table}[t!]
\centering
\fontsize{8pt}{13pt}\selectfont
\resizebox{\columnwidth}{!}{%
\begin{tabular}{lccccc}
\hline
\textbf{Model} & \textbf{HPQA} & \textbf{2Wiki} & \textbf{Musi.} & \textbf{\( \overline{\textbf{EM}} \)} & \textbf{\( \overline{\textbf{tools}} \)} \\
\hline
SearchR1-PPO & 30.00 & 31.50 & 10.00 & 23.83 & 4.51 \\
OTC-GRPO\textdagger & 36.60 & 31.10 & 13.00 & 26.90 & \textbf{1.05} \\
OTC-PPO\textdagger & 38.30 & 36.30 & 15.20 & 29.93 & 1.76 \\
StepSearch\textdagger & 38.60 & 36.60 & \textbf{22.60} & 32.60 & - \\
\hline
$\text{TEPO}_{\text{sparse}}$ & 42.75 & \textbf{54.12} & 19.62 & \textbf{38.83} & 2.24 \\
$\text{TEPO}_{\text{dense}}$ & \textbf{43.60} & 51.90 & 20.60 & 38.70 & 5.77 \\
\hline
\end{tabular}%
}
\caption{Results on five evaluation runs. \textdagger indicates that the results are directly cited from the paper. All baselines are based on Qwen2.5-7B-Instruct. \textbf{$\overline{\text{EM}}$} denotes the average exact match, included to ensure comparability with prior work.}
\label{extra_reasoning_results}
\end{table}

\subsection{Baselines}
We compare against two groups of baselines. 
(i) \textbf{Algorithm-level baselines:} methods trained under the same pipeline as ours (same SFT stage, training data, and hyperparameters unless stated otherwise), so that differences can be attributed to the optimization objective and reward design. 
This group includes GRPO, a standard group-based policy optimization method, and two entropy-driven RL objectives, ARPO~\cite{dong2025arpo} and AEPO~\cite{dong2025aepo}.
(ii) \textbf{Recent TIR baselines:} recent RL methods that also use wiki-18 as the knowledge source but may adopt different training recipes. 
This group includes SearchR1~\cite{jin2025searchr1trainingllmsreason}, StepSearch~\cite{wang2025stepsearch}, and OTC-PO~\cite{wang2025otc}. To account for variance, we run each evaluation five times and report the mean and standard deviation.

\begin{table*}[htbp]
\fontsize{7pt}{12pt}\selectfont
\centering
\caption{Overall performance on deep search tasks. The best results are in \textbf{bold}, and the second-best results are \underline{underlined}. \textbf{$\overline{\text{acc}}$} represents the average accuracy, and \textbf{$\overline{\text{tools}}$} represents the average number of tool calls, both calculated across all three datasets.}
\label{tab:deepsearch_results}
\begin{tabularx}{\textwidth}{l*{6}{>{\centering\arraybackslash}X} *{2}{>{\centering\arraybackslash}X}}
\toprule
\multirow{2}{*}{\textbf{Model}} & \multicolumn{2}{c}{\textbf{webw.}} & \multicolumn{2}{c}{\textbf{hle}} & \multicolumn{2}{c}{\textbf{gaia}} & \multirow{2}{*}{$\overline{\text{\textbf{acc}}}$} & \multirow{2}{*}{$\overline{\text{\textbf{tools}}}$} \\
\cmidrule(lr){2-3} \cmidrule(lr){4-5} \cmidrule(lr){6-7}
& \textbf{acc} & \textbf{tools} & \textbf{acc} & \textbf{tools} & \textbf{acc} & \textbf{tools} & & \\
\midrule
\textbf{Qwen3-8B} & 1.00 ± 0.84 & \textbf{1.67 ± 0.17} & 5.76 ± 0.77 & \underline{0.34 ± 0.03} & 12.04 ± 1.58 & \underline{1.41 ± 0.17} & 6.27 & \textbf{1.14} \\
\textbf{\hspace{0.8em}+ SFT} & 5.30 ± 1.03 & 7.70 ± 0.34 & 6.36 ± 0.74 & 1.68 ± 0.06 & 15.53 ± 2.38 & 7.56 ± 0.39 & 9.06 & 5.65 \\
\textbf{\hspace{0.8em}+ GRPO} & \underline{5.80 ± 1.17} & 2.87 ± 0.20 & 6.36 ± 0.73 & 0.57 ± 0.03 & 16.12 ± 2.50 & 1.85 ± 0.10 & 9.43 & 1.77 \\
\textbf{\hspace{0.8em}+ ARPO} & 5.40 ± 1.28 & 6.75 ± 0.26 & 6.16 ± 0.34 & 1.04 ± 0.07 & 16.89 ± 2.00 & 6.11 ± 0.28 & 9.48 & 4.63 \\
\textbf{\hspace{0.8em}+ AEPO} & \textbf{6.00 ± 1.30} & 5.06 ± 0.05 & \textbf{6.76 ± 0.92} & 0.92 ± 0.06 & 16.31 ± 0.95 & 3.64 ± 0.22 & \underline{9.69} & 3.21 \\
\textbf{\hspace{0.8em}+ $\text{TEPO}_{\text{sparse}}$} & 4.90 ± 0.73 & \underline{2.28 ± 0.20} & 6.40 ± 0.42 & \textbf{0.29 ± 0.02} & \textbf{17.67 ± 2.16} & \textbf{1.12 ± 0.22} & 9.66 & \underline{1.23} \\
\textbf{\hspace{0.8em}+ $\text{TEPO}_{\text{dense}}$} & 5.60 ± 1.11 & 2.90 ± 0.16 & \underline{6.72 ± 0.48} & 0.63 ± 0.02 & \underline{17.28 ± 1.13} & 2.00 ± 0.09 & \textbf{9.87} & 1.84 \\
\bottomrule
\end{tabularx}
\end{table*}

\subsection{Main Results}
\paragraph{Results on Reasoning Tasks.} The results are presented in Table~\ref{math and search results}. $\text{TEPO}_{\text{sparse}}$ yields a substantial improvement in tool-use efficiency across different models, reducing tool calls by 72.07\% compared to the average of baselines while still showing comparable performance. This is expected because $\text{TEPO}_{\text{sparse}}$ employs an outcome-level reward in which the number of tool calls appears as a denominator term, thereby providing a global training signal that encourages the agent to achieve correct outcomes with fewer calls. 

In contrast, $\text{TEPO}_{\text{dense}}$ outperforms all baselines in reasoning performance with an average increase of 22.27\%, highlighting the advantage of its process-level reward: by assigning fine-grained credit to individual tool calls, it better shapes step-wise tool-use decisions.

\begin{figure}[t!]
  \centering
  \includegraphics[width=\columnwidth]{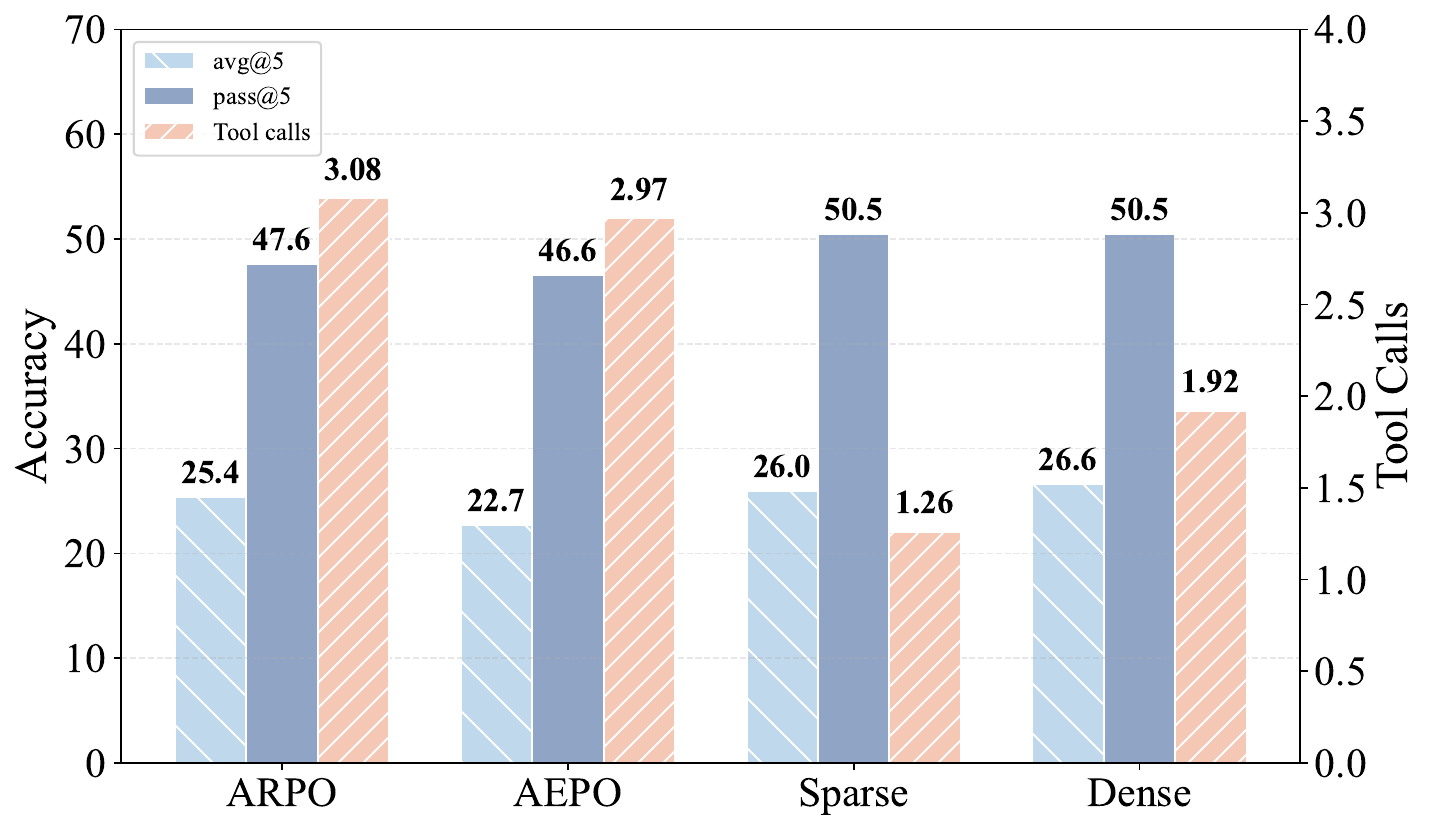}
  \caption{Results on GAIA dataset evaluated with Qwen3-8B using Bing Search API as search tool, including avg@5, pass@5, and average tool calls.}
  \label{fig:bing}
\end{figure}

Additionally, the results in Table~\ref{extra_reasoning_results} show that $\text{TEPO}_{\text{sparse}}$ and $\text{TEPO}_{\text{dense}}$ exceed the performance of several recent baselines. The OTC methods, which focus on tool-use efficiency, show higher efficiency than $\text{TEPO}_{\text{sparse}}$, while the latter achieves a performance boost. Overall, these results demonstrate that our proposed entropy-guided algorithms are both effective and robust.

\paragraph{Results on Deep Search Tasks.} The main experimental results are shown in Table~\ref{tab:deepsearch_results}. Additionally, we conducted extra experiments using the Bing Search API as a retrieval tool, which better supports deep search tasks. The results are shown in Figure~\ref{fig:bing}, and the conclusions remain consistent: $\text{TEPO}_{\text{sparse}}$ demonstrates stronger tool-use efficiency, while $\text{TEPO}_{\text{dense}}$ exhibits superior reasoning performance.

\section{Analysis}

\begin{figure}[t!]
  \centering
  \includegraphics[width=\columnwidth]{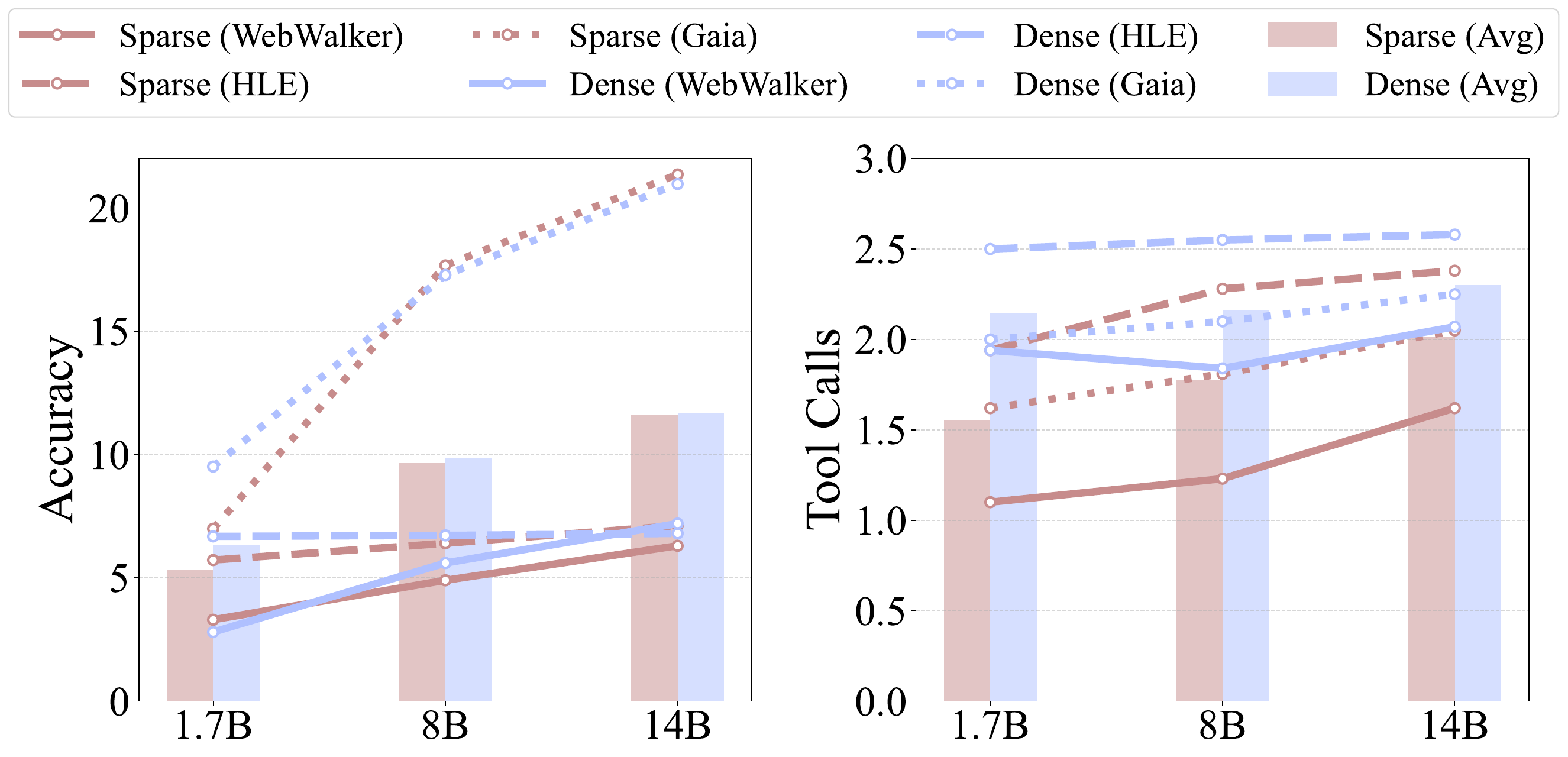}
  \caption{Results on Deep Search Tasks evaluated with different sizes of Qwen3 models using wiki-18 as search corpus, including avg@5 and average tool calls.}
  \label{fig:scaling}
\end{figure}

\subsection{Scaling Analysis.}
To further assess the scalability of our methods, we conducted an analysis across different model sizes, as shown in Figure \ref{fig:scaling}. Our results show that both $\text{TEPO}_{\text{sparse}}$ and $\text{TEPO}_{\text{dense}}$ exhibit strong scalability. As the model size increases from 1.7B to 14B, performance consistently improves for both methods, following a clear scaling law. However, the scaling law is less pronounced on the HLE dataset, suggesting that the knowledge involved may exceed the capabilities of both the model and the wiki-18 knowledge base. This indicates the need for more powerful knowledge retrieval tools to further enhance performance. Notably, tool-use efficiency remains relatively stable across different scales, demonstrating the effectiveness in optimizing tool calls without compromising efficiency.

\subsection{Training Efficiency Analysis}
\begin{figure}[t!]
  \centering
  \includegraphics[width=\linewidth]{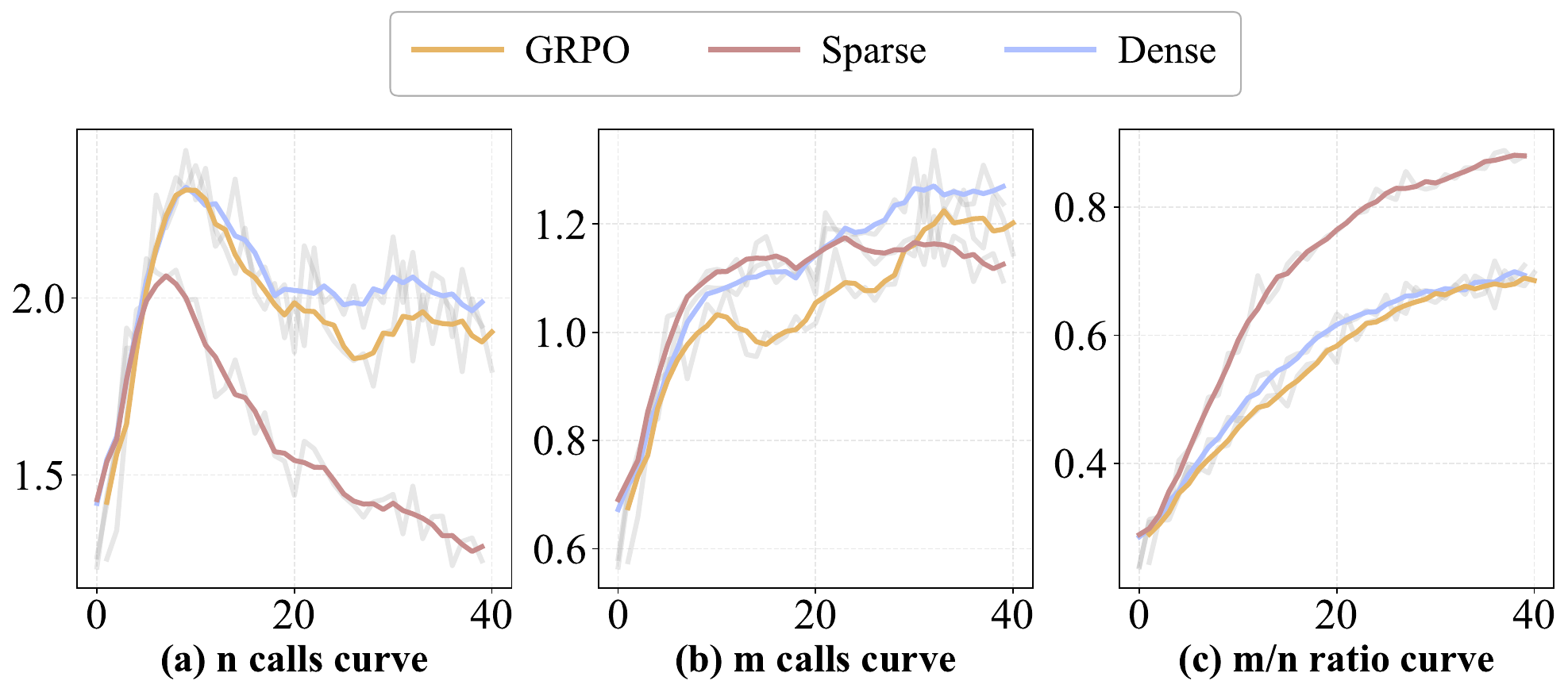}
    \caption{Visualization of training dynamics, showing (a) \( n \)  curve, (b) \( m \) curve and (c) \( m/n \) ratio curve. Here, \( n \) is the total number of tool calls, and \( m \) is the number of tool calls inducing entropy reduction.}
  \label{fig:tool_calls_qwen3}
\end{figure}

\paragraph{Tool-Call Efficiency Analysis.}  
Figure~\ref{fig:tool_calls_qwen3} (a) shows the tool invocation curves during the training process. All three algorithms exhibit an initial increase in tool calls followed by a gradual decrease. Notably, $\text{TEPO}_{\text{sparse}}$ shows the fastest reduction in tool calls, which is consistent with the earlier finding that the $\text{TEPO}_{\text{sparse}}$ method demonstrates higher tool-call efficiency.

\paragraph{Entropy-Reducing Tools Analysis.} 
We further investigated the relationship between the number of tool calls that induce entropy reduction across different algorithms during training. We define \( n \) as the total number of tool calls and \( m \) as the number of tool calls inducing entropy reduction, with the ratio \( m/n \) reflecting the proportion of entropy-reducing tool calls. The results are shown in Figure~\ref{fig:tool_calls_qwen3} (b), which presents the \( m \) calls curve, and Figure~\ref{fig:tool_calls_qwen3} (c), which shows the \( m/n \) ratio curve. 

As can be seen from the figures, $\text{TEPO}_{\text{sparse}}$ exhibits a rapid increase in the \( m/n \) curve. This is largely driven by its sparse reward design, which uses \( m/n \) as a scaling factor: optimizing this ratio can be achieved either by increasing \( m \) or by reducing \( n \). As a result, $\text{TEPO}_{\text{sparse}}$ improves tool-use efficiency while also raising the proportion of tool calls that induce entropy decrease. In contrast, $\text{TEPO}_{\text{dense}}$ shows a larger growth in \( m \), which is directly attributable to its fine-grained reward design. By providing dense tool-level process rewards, $\text{TEPO}_{\text{dense}}$ explicitly encourages each tool call to be entropy-reducing, leading to more consistently effective tool use and ultimately better reasoning performance.

\begin{table}[t]
\centering
\small
\sisetup{detect-all, table-number-alignment=center}

\resizebox{\columnwidth}{!}{
\begin{tabular}{@{\extracolsep{\fill}} ccccccc}
\toprule
\textbf{Score} & \textbf{Num.} & \textbf{Webwalker} & \textbf{HLE} & \textbf{Gaia} & \textbf{Avg} \\
\midrule
\rowcolor{gray!12}
\multicolumn{6}{c}{\textbf{\textit{Qwen3-8B-SFT}}} \\
\midrule
0 & 13002 & 16.33 & -18.22 & -5.90 & -2.60 \\
1 & 2122  & -20.81 & -27.72 & -26.78 & -25.10 \\
\midrule
\rowcolor{gray!12}
\multicolumn{6}{c}{\textbf{\textit{Qwen3-8B-Sparse}}} \\
\midrule
0 & 2814  & -0.71 & -65.23 & -55.57 & -40.50 \\
1 & 551   & -53.09 & -71.85 & -75.59 & -66.84 \\
\midrule
\rowcolor{gray!12}
\multicolumn{6}{c}{\textbf{\textit{Qwen3-8B-Dense}}} \\
\midrule
0 & 4435  & 16.71 & -39.75 & -7.69 & -10.24 \\
1 & 1008  & -17.34 & -54.40 & -34.74 & -35.49 \\
\bottomrule
\end{tabular}
}
\caption{Delta entropy experiment results: We report the average delta segment entropy ($\Delta H_k$) of SFT, Sparse, and Dense models on deep search tasks across three datasets, aggregated by tool scores. For readability, $\Delta H_k$ values are scaled by a factor of $10^3$.}
\label{delta entropy analysis}
\vspace{-10pt}
\end{table}

\subsection{Delta Entropy Analysis.}
To further explore the impact of entropy reduction signals on tool usage, we examined the entropy changes induced by tool calls under different training methods. Delta entropy analysis was conducted on the three datasets from the deep search tasks, focusing on the SFT model as well as the $\text{TEPO}_{\text{sparse}}$ and $\text{TEPO}_{\text{dense}}$ models.

Specifically, we conducted five evaluation runs on these datasets, and following the experimental method outlined in section~\ref{pilot experiment}, we computed the average delta segment entropy ($\Delta H_k$) for both high-quality and low-quality tool calls. The results are shown in Table~\ref{delta entropy analysis}. 

After training with $\text{TEPO}_{\text{sparse}}$ and $\text{TEPO}_{\text{dense}}$, both low-quality and high-quality tools exhibit a stronger tendency for entropy reduction compared to the SFT model, with $\text{TEPO}_{\text{sparse}}$ showing a more pronounced entropy reduction. However, the proportion of high-quality tools in $\text{TEPO}_{\text{sparse}}$ (16.4\%) remains lower than that in $\text{TEPO}_{\text{dense}}$ (18.5\%).

The above results provide complementary evidence for the effectiveness of both proposed algorithms from another perspective. $\text{TEPO}_{\text{sparse}}$ is built upon an outcome-level reward and thus optimizes tool-use behavior at a global level, which manifests as fewer total tool calls and more entropy-reducing tool calls. On the other hand, $\text{TEPO}_{\text{dense}}$ adopts a fine-grained reward design that encourages entropy-reducing tool calls at each step, also leading to a stronger entropy-reduction tendency than the SFT model. These findings suggest that entropy reduction can serve as an effective supervision signal for both $\text{TEPO}_{\text{sparse}}$ and $\text{TEPO}_{\text{dense}}$.

\section{Related Work}
\paragraph{Agentic RL for Tool-Integrated LLMs. } Tool-integrated LLM agents require policies that interleave reasoning with environment actions (e.g., search, API calls), shifting the learning problem from single-shot generation to long-horizon interaction~\cite{yao2022react,nakano2021webgpt,schick2023toolformer,qin2023toolllm,chen2023chatcot, gou2023critic}. With reinforcement learning becoming a popular paradigm for training, recent work has utilized reinforcement learning to train models to use tools such as Python interpreters or search tools, enabling them to perform complex, multi-step, long-horizon reasoning tasks \cite{qian2025toolrl, gao2025beyond, li2025torl, he2025webseer, wang2025erase, zhang2025tool}.

However, in long trajectories, agents may invoke tools excessively or inappropriately, increasing computation cost and derailing the reasoning process~\cite{qian2025smart}. Therefore, recent work has started to focus on the behavior of models when using tools \cite{wang2025toward, jin2025beneficial}. For example, \citet{wang2025otc} explores how to train models to learn the optimal number of tool calls, improving tool invocation efficiency. Similarly, \citet{li2025encouraging} introduces a reward signal based on tool-use completeness, which enhances the model's ability to call tools effectively. 

Although recent studies have explored process rewards, such as \cite{zhang2025rlvmr, feng2025group}, how to provide step-level supervision signals for tool invocation behavior without a process reward model or handcrafted rules remains an open and valuable research question. The entropy reduction pattern serves as an effective lightweight supervision signal, enabling the model to learn what constitutes good tool-use behavior during training.

\paragraph{Entropy-Based Signals for Agentic RL.} Entropy and information-theoretic metrics provide intrinsic uncertainty signals~\cite{li2025confidence, sharma2025think, stoisser2025towards} that guide agent behavior. \citet{yong2025think} investigates step-wise information gain and adaptive termination for efficient reasoning. In agentic RL, \citet{dong2025arpo} and \citet{dong2025aepo} primarily leverage entropy for exploration and training stability, allocating more branching and sampling around high-uncertainty steps. Similarly, \citet{cheng2025reasoning} demonstrates that higher entropy in reasoning correlates with exploratory behaviors, suggesting that entropy can drive deeper, more comprehensive reasoning chains. 

While these studies use entropy to encourage exploration during high-uncertainty steps, this paper takes a different approach by directly rewarding tool calls that reduce entropy. Instead of promoting exploration, we reinforce behaviors that decrease uncertainty through tool use, with a reduction in entropy signaling positive information gain and improved reasoning performance.

\section{Conclusion}
In this work, we first delve into the relationship between tool usage and entropy. Through pilot experiments, we found that high-quality tool calls are often accompanied by entropy reduction. Building on this finding, we propose using entropy reduction as a supervisory signal and introduce two distinct reward strategies, each tailored to optimize tool-use behavior in different contexts. $\text{TEPO}_{\text{sparse}}$ incorporates the proportion of entropy-reducing tool calls into final-answer correctness, reducing overall tool usage while improving efficiency and increasing the proportion of entropy-reducing tools for better reasoning performance. In contrast, $\text{TEPO}_{\text{dense}}$ uses entropy reduction as a process-level supervision signal, guiding the model to recognize and optimize good tool usage behavior during training. Extensive experiments on several datasets validate the effectiveness of both methods. $\text{TEPO}_{\text{sparse}}$ prioritizes efficiency, while $\text{TEPO}_{\text{dense}}$ focuses on performance. This trade-off enables flexible tool usage strategies for different tasks. Our findings demonstrate that entropy reduction can serve as a powerful signal in reinforcement learning, paving the way for future research to refine entropy-based reward mechanisms and train more adaptive agents for complex environments.

\section*{Limitations}
Although the entropy reduction-based reward design proposed in this paper demonstrates promising results across various reasoning tasks, there are still some limitations.
Firstly, while we validated the scalability from 1.7B to 14B models in our experiments, due to computational resource limitations, experiments on larger model sizes were not conducted. Future work could involve testing on larger models to further assess the performance of the proposed method across different model scales. Secondly, the primary experiments in this paper used the wiki-18 search tool, chosen for its stability and reproducibility in controlled settings. However, more challenging datasets and recent knowledge domains would benefit from the use of real-time search APIs, such as Bing or Google. While we conducted an experiment using the Bing Search API on the GAIA dataset, due to API cost and stability considerations, we did not test additional search APIs. In future work, we plan to expand the experimentation to include other real-time search APIs, ensuring broader validation and effectiveness in scenarios requiring up-to-date knowledge.



\bibliography{custom}

\appendix

\clearpage
\section{Details for Entropy-Based Pilot Experiments}
\label{appendix:A}

\subsection{Domain Partition and Datasets}
We follow the same domain partition as in ARPO~\cite{dong2025arpo} and AEPO~\cite{dong2025aepo}. 
The Mathematical Reasoning domain includes AIME2024 and AIME2025, consisting of competition-level problems that require multi-step numerical and symbolic reasoning. 
The Knowledge-Intensive Reasoning domain includes open-domain multi-hop QA benchmarks that benefit from external search, including HotpotQA~\cite{yang2018hotpotqa}, 2WikiMultihopQA~\cite{ho2020constructing}, and Musique~\cite{trivedi2022musique}. 
The Deep Information Searching domain includes web-search benchmarks that require iterative retrieval and long-horizon decision making, including GAIA~\cite{mialon2023gaia}, WebWalker~\cite{wu2025webwalker}, and HLE~\cite{phan2025humanity}.

\subsection{LLM-as-Judge for Tool-Call Quality}
During evaluation, we adopt GPT-4o-mini as an LLM-as-judge to assess the quality of each tool call.
For every tool call at step $k$, the judge is provided with: (i) the original question, (ii) the context preceding the call, (iii) the tool query, (iv) the tool result returned by the executor $E$, and (v) the subsequent response segment after observing the tool result. 
The judge assigns a binary tool score $y_k \in \{0,1\}$, where $y_k=1$ indicates a high-quality tool call that provides relevant and useful information for solving the task, and $y_k=0$ indicates a low-quality call, such as malformed queries, failed executions, or irrelevant results. 
The full prompt used by the LLM-as-judge is provided in Appendix~\ref{appendix:judge_prompt}.

\subsection{Evaluation Details and Entropy Statistics}
For each domain and each SFT model, we run five independent evaluation trials.
For each trial, we collect all tool calls made by the model and compute entropy-based statistics for each tool call.

\paragraph{Per-call entropy change.}
Following the entropy indicator defined in Sec.~\ref{formalization of delta entropy}, we compute the delta segment entropy $\Delta H_k$ for each tool call at step $k$, measuring the change of uncertainty in the subsequent response segment after the tool interaction.
A negative $\Delta H_k$ indicates entropy reduction (i.e., decreased uncertainty), while a positive value indicates increased uncertainty.

\paragraph{Per-call entropy change ratio.}
We also compute the delta segment entropy ratio $\Delta H_k^{\text{ratio}}$ for each tool call that induces an entropy decrease, which normalizes the entropy change by the corresponding segment scale as defined in Sec.~\ref{formalization of delta entropy}. 
This ratio enables comparisons of entropy dynamics across domains and models with different segment lengths and distributions.

\subsection{Judge Prompt}
\label{appendix:judge_prompt}
We include the prompt template used by GPT-4o-mini for assigning tool-call quality scores.

\begin{figure}[H]
\centering
\begin{tcolorbox}[title=Utility Scoring Prompt]
\textbf{You are an expert evaluator assessing the quality of tool calls in an AI agent's problem-solving process.}

Original Question: \texttt{\{question\}} \\
Context previous to Tool Call: \texttt{\{previous\_context\}} \\
Search Query: \texttt{\{search\_query\}} \\
Tool Result: \texttt{\{tool\_result\}} \\
Agent Response subsequent to Tool Result: \texttt{\{subsequent\_response\}}

For the tool call, assign a score of 0 or 1. The criteria for evaluation are as follows:
\begin{itemize}
  \item If the tool call itself fails, or if the search query derived from the previous context is unreasonable and cannot help answer the question, or if the tool result is irrelevant, unhelpful, or incorrect, assign a score of 0.
  \item If the tool call result provides help in reasoning about the problem or advances the reasoning process, assign a score of 1.
\end{itemize}

Respond with only the score (0 or 1) followed by a brief explanation classification: If the score is 0, give the classification in \texttt{[unreasonable search query / search api failed / irrelevant or incorrect search result]}. If the score is 1, no need to provide explanation.

Format: \texttt{SCORE: <0 or 1>: <explanation classification>}
\end{tcolorbox}

\caption{LLM-as-Judge prompt for scoring tool quality.}
\label{fig:utility_scoring_prompt}
\end{figure}

\section{Details for Experiment Settings}
\label{appendix:B}

\subsection{Dataset Details}
\label{appendix:B.1}
The datasets used in our experiments span three distinct domains, each focusing on different aspects of reasoning abilities. Mathematical Reasoning evaluates the model’s ability to solve competition-level problems that require multi-step numerical and symbolic reasoning. Knowledge-Intensive Reasoning tests the model's capacity for multi-hop question answering, where external search is leveraged to retrieve relevant information. Finally, Deep Information Searching assesses the model's performance in web-search tasks that require iterative retrieval and long-horizon decision-making, where the model needs to make decisions based on continuously evolving information.

Dataset sizes are shown in Table~\ref{details of datasets}. Notably, in order to facilitate the evaluation, only the first 500 questions of the HLE dataset were selected. All tests involving the use of the HLE dataset in this paper were conducted with fair comparisons, where the results are based on the average of five evaluation runs on the first 500 questions.

\begin{table}[t!]
\centering
\fontsize{8pt}{13pt}\selectfont
\resizebox{\columnwidth}{!}{%
\begin{tabular}{ccc}
\hline
\textbf{Datasets} & \textbf{Number of Questions} & \textbf{Domain} \\
\hline
AIME2024 & 30 & Math Reasoning \\
AIME2025 & 30 & Math Reasoning \\
\hline
HotpotQA & 200 & Knowledge Intensive Reasoning \\
2wiki & 200 & Knowledge Intensive Reasoning \\
Musique & 200 & Knowledge Intensive Reasoning \\
\hline
GAIA & 103 & Deep Information Searching \\
WebWalker & 200 & Deep Information Searching \\
HLE & 500 & Deep Information Searching \\
\hline
\end{tabular}%
}
\caption{Details of datasets used in the main experiment.}
\label{details of datasets}
\end{table}

\subsection{Implementation Details}
\label{implementation details}
\noindent\textbf{SFT Training Stage:} We initialize the tool-use behaviors via supervised fine-tuning using the LLaMAFactory framework. The SFT training corpus primarily sourced from the Tool-Star~\cite{dong2025toolstar} open-source dataset, along with some STILL dataset, totaling 54K data samples.

\noindent\textbf{RL Training Stage:} Building on the SFT training, we refine the model using reinforcement learning, selecting distinct RL training sets for different test domains, as in ARPO/AEPO. For reasoning tasks, we use 10K open-source RL samples from Tool-Star, primarily evaluating the performance on Qwen2.5 and Llama3.1 models. For deep search tasks, we use a 1K RL training set, mainly assessing the model's performance on Qwen3.

In our implementation, we adopt VERL~\cite{verl} framework to train the models. To stabilize reinforcement learning training, we exclude the tool-call results from loss computation to avoid bias, and we set the KL divergence coefficient to zero. We configure the rollout number to 8 to balance sample efficiency and training stability. During reinforcement learning, we train with a batch size of 128, a PPO minibatch size of 16, and a context window of 20K tokens. Training epoches are set to 2 and 5 separately for Reasonsing tasks and Deep Search tasks. All experiments are conducted on four NVIDIA H200 GPUs. In the evaluation process, we adopt the VLLM framework with the inference parameters set as follows: top-p, temperature, and top-k are set to 0.95, 0.2, and 20, respectively. After extracting the answers, we use GPT-4o-mini to evaluate their correctness.

\section{Failed Cases Analysis.}
\label{appendix:C}
\begin{figure}[t!]
  \centering
  \includegraphics[width=\columnwidth]{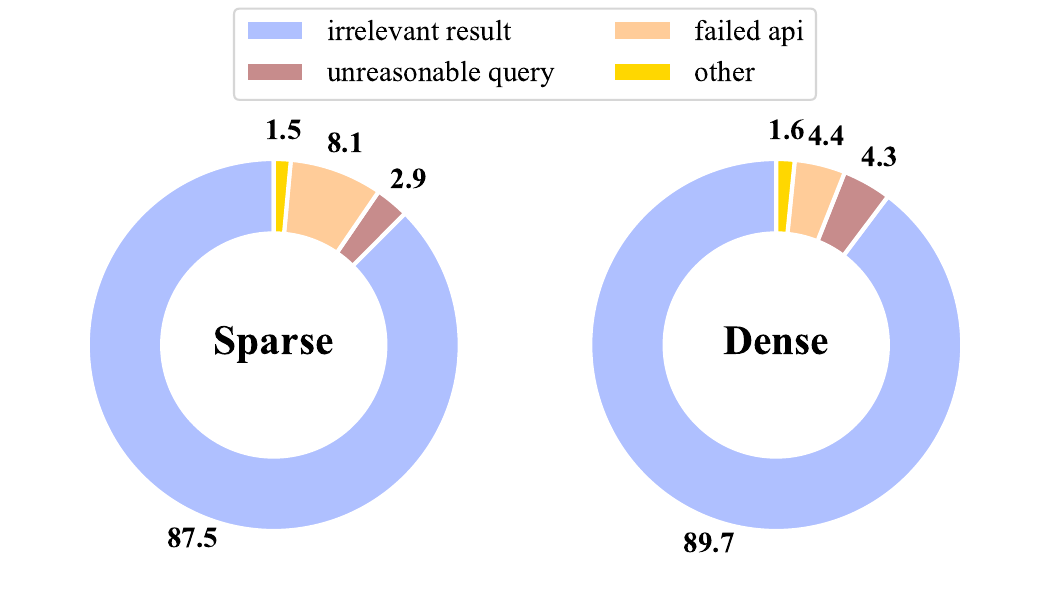}
  \caption{Results on the GAIA dataset using the Bing search API. The reasons for each tool call labeled with a score of 0 are summarized. All values in the figure are presented as percentages.}
  \label{fig:bad_case}
\end{figure}

As shown in Figure~\ref{fig:bad_case}, most low-quality tool calls were due to the inability to find relevant results. This suggests that while the model is trained to generate tool queries that aim to retrieve critical information for reasoning, these queries may not always be fully effective when interacting with external search APIs. Although the queries are often direct, they may not be optimally structured to retrieve relevant data. This highlights the importance of future research, which should not only focus on teaching models to use tools effectively but also on enhancing their ability to craft and refine search queries. Improving the query construction process can lead to more precise interactions with external tools, ensuring that the model retrieves the most relevant and accurate information.
\end{document}